# Minimizing Perceived Image Quality Loss Through Adversarial Attack Scoping

Kostiantyn Khabarlak    Larysa Koriashkina

National Technical University Dnipro Polytechnic, Ukraine

**ABSTRACT**

Neural networks are now actively being used for computer vision tasks in security critical areas such as robotics, face recognition, autonomous vehicles yet their safety is under question after the discovery of adversarial attacks.

In this paper we develop simplified adversarial attack algorithms based on a scoping idea, which enables execution of fast adversarial attacks that minimize structural image quality (SSIM) loss, allows performing efficient transfer attacks with low target inference network call count and opens a possibility of an attack using pen-only drawings on a paper for the MNIST handwritten digit dataset.

The presented adversarial attack analysis and the idea of attack scoping can be easily expanded to different datasets, thus making the paper's results applicable to a wide range of practical tasks.

## 1 INTRODUCTION

Year of 2012 has seen a paradigm shift from human-generated feature extractors for computer vision applications, in particular image classification, to self-learnt classification systems named Neural Networks, when AlexNet won ImageNet Large Scale Visual Recognition Challenge with an error rate significantly lower than that of competing solutions. Neural networks are being learnt by feeding large amounts of data and changing its inner representation to lower error using Back Propagation algorithm. That way, researchers have attempted to copy the way a child is being learnt. Yet one piece is missing – a way of interrogating and validating what exactly the network has learnt. The initial attempts in that direction have been made by looking at statistical distributions [2], finding network components sensitive to a specific feature [3] or generating an image causing a maximum activation of a specific network unit [4]. However, we still have no full understanding of the neural network inference process.

The ongoing research in a direction of neural network inference understanding has led to a discovery coined as Adversarial Attack [3]. By adding a slight human-imperceptible perturbation to an input image, it has been made possible to fool the network and force it to misclassify the image as an instance of another class. Such an attack has been further divided into 2 subgroups: 1) the non-targeted attack, when the new inference class is of no importance unless it's different from the source; 2) the targeted attack, when we specify the target class we want the network to predict after the attack for a given image.

In this work we: 1) provide an insight into the attack procedure; 2) develop a new set of fast targeted and non-targeted algorithms that minimize structural image quality loss; 3) compare the algorithms against the ones previously developed and explore new attack applications by utilizing our algorithm's features.

## 2 REVIEW OF THE LITERATURE

While neural networks are considered to be an efficient solution to many problems including, but not limited to object detection in images, speech recognition, natural language processing, their inner workings are not entirely understood. One of the first attempts to better understand the neural network inference apparatus for image classification [3] has led to an interesting neural networks property called by the authors "Adversarial attack". The idea was to investigate how input modification affects neural network's output. As it has been discovered, a slight change to the input is enough for the neural network to misclassify the image. However, it is highly unlikely to achieve such an effect via random search over input modifications, the change must be directed. The first algorithm performing such a directed search is Box-constrained L-BFGS [3]. Later, a faster and simpler algorithm FGSM (Fast Gradient Step Method) has been developed, where first-order loss function derivative approximation has been used to generate adversarial examples [6]. The attack is performed in a single step, which in practice leads to either overly perturbed images or images that are failing to trick the network. I-FGM is an iterative method, which performs an attack until it succeeds. Another simple enhancement over FGSM is made, that is instead of using a sign of gradient of loss function, the gradient itself can be used, thus leading to more efficient and less noisy adversarial images [7]. Another captivating property of adversarial examples is their transferability, i.e. image can be generated for one network and blindly transferred attack a different network with a success. Simple and fast algorithms are still under active development, by using gradient with momentum it has been possible to improve adversarial image generation and improve their transferability to other networks [8].

Another group of algorithms is those that are trying to generate adversarial images with particular properties. In [9] Jacobian-Based Saliency Map algorithm has been introduced. By using forward derivative and building Jacobian matrix of the function learned by the neural network, authors try to find points that will change the output of the network the most. By modifying them one at a time from the most important to the least, they essentially target $L_1$ quality metric, that is the number of pixels modified during the attack. An algorithm that makes it possible to generate a universal noise mask once and then reapply it to other images to misclassify them is presented in [10]. Although this algorithm allows us to generate new adversarial images for free without an expensive computation, the images generated do not pursue quality goals and only non-targeted attack can be performed.

It has been shown that adversarial attacks can be conducted in real-world too. Noisy images printed using color printer can make neural network perceiving through mobile-phone camera to misclassify images [12]. Autonomous vehicles are also vulnerable to attacks as noisy stickers can be printed and attached to road signs, causing a severe misclassification of "Stop" sign as a "Speed Limit". The attack is robust to distance and viewing angle change, thus potentially causing a road accident [11]. Face recognition systems are in effect as special googles can be developed for the human to be misclassified as another person or to evade detection all together [13].

An active research in the domain of adversarial attack defense and detection is being conducted. An exhaustive attack and defense catalog can be found in the works [16, 17]. Notwithstanding a diligent investment is this area, no security measure known to date has survived newer generation of adversarial attacks. In [19] it has been shown that it is possible to transfer images to secured networks or networks with an unknown architecture or trained on an unknown dataset (classes of the dataset are still known). In [15] authors have developed C&W adversarial attack algorithm, that optimizing a special loss function is able to generate more robust attack images, which allowed them to generate more efficient adversarial examples compared to the algorithms known to date and to crack Defensive Distillation algorithm. C&W algorithm targets purely white-box attack algorithm (transfer to a different network is still possible). ZOO algorithm presented in [14] is an adaptation of C&W attack for a black-box scenario. Though is more effective, it is very slow. Finally, by adjusting C&W algorithm loss function, it has been shown that it is possible to break 10 more adversarial attack detection algorithms [20].

Many of the above-described adversarial attack and defense methods can be found in an opensource collection known as cleverhans library [14].

An important point for further adversarial attack defense research is to better understand the precursors of the attack. Thus, an attempt to present an efficient, fast and simple algorithm has been made in this paper. The key goals are: 1) the algorithm's attack flow should be easily understood and implemented; 2) the algorithm should be fast, so that generalized algorithm performance statistics could be collected on large arrays of images; 3) the algorithm should enable new types of neural network attacks.

## 3 MATERIALS AND METHODS

### 3.1 Problem statement

Let $I$ – feature space dimensionality, neuron count in the network's input layer; $K$ – number of classes, neuron count in the output layer; $x$ – network's input vector, image's pixel brightness vector; $y$ – unit vector, which defines object attribution to one class or another; $z$ – network's output vector.

Given a training set
$X = \{x^{(j)}, y^{(j)}\}_{j=1}^{J}: x^{(j)} \in R^I; y^{(j)} \in R^K$,
$y_k^{(j)} = 0 \vee 1 \ \forall j, \sum_{k=1}^{K} y_k^{(j)} = 1 \ \forall j$.

Let $W: [I \times K]$ and $b = (b_1, b_2, \ldots, b_K)$ network's weight matrix and biases vector, which minimize function

$$G(W,b) = \frac{1}{M} \sum_{m=1}^{M} \gamma(W,b; z^{(m)}, y^{(m)}) =$$
$$= -\frac{1}{M} \sum_{m=1}^{M} \sum_{k=1}^{K} y_k^{(m)} \log \left( \sigma_k(x^{(m)} \cdot W + b) \right),$$

where
$\gamma(W,b; z, y) = -\sum_{k=1}^{K} y_k \log(z_k)$,
$z = \sigma(x \cdot W + b)$,
$\sigma(z) = (\sigma_1(z), \sigma_2(z), \ldots, \sigma_K(z)), \sigma_k(z) = \frac{e^{z_k}}{\sum_{j=1}^{K} e^{z_j}}$,

$M$ – training batch size, used to define number of images used in a single optimization method step during training; $\sigma_k(\cdot)$ – output layer's $k^{th}$ neuron softmax activation function, which computes layer output by its input.

As is known, *softmax* serves a goal of transforming an arbitrary real-valued vector into a probability distribution of the inferred classed. Cross-entropy loss $\gamma(W,b; z, y)$ defines an error metric between computed outputs $y$ and desired $z$.

We denote $S$ – source class, $T$ – target class (desired result), $\hat{z}$ – desired (target) neural network's output.

1. Find such a perturbation $\Delta x \in R^I$, which for a given value $T \in \{1, 2, \ldots, K\}$ and deliberately chosen $x$, such that $z_S = \max_{k=1,K} z_k$, $S \neq T$,

$$\hat{z}_T = \max_{k=1,K} \hat{z}_k,$$

where $\hat{z} = \sigma((x + \Delta x) \cdot W + b)$.

2. Find such a perturbation $\Delta x \in R^I$, which for a deliberately chosen $x$, such that $z_S = \max_{k=1,K} z_k$, $l \in \{1, 2, \ldots, K\}$ can be found that meets a condition of $\hat{z}_l > \hat{z}_S$.

### 3.2 Generating Adversarial Examples

We consider the MNIST dataset because of its small size and ability to make accurate predictions even using simple neural networks. The dataset consists of normalized grayscale handwritten digit images in range $0 - 9$ of size $28 \times 28$. The dataset has been split into training subset with 60,000 examples and testing subset 10,000. Each subset contains samples of digits handwritten by distinct people.

For the neural network training and inference process each image is unrolled into single-dimensional vector. Pixel intensities are normalized into $[0,1]$ range by dividing its values by 255, then 0 stands for black pixels, 1 for the white ones.

A single-layer neural network (logistic regression) has been chosen as a target network architecture. It is built with an input layer of $I = 784$ neurons (each input pixel is considered as a separate input feature) and $K = 10$ in the output layer (by the number of classes). The logistic regression's key advantage, which will be used further down to build an attack algorithm, is interpretability of a weight matrix $W_{ik}$ as of an importance or a contribution of $i^{th}$ image pixel towards $k^{th}$ class classification. Precisely, if

$W_{ik} > 0$, it is expected that an increase of a corresponding pixel brightness by a some $\delta > 0$ will lead to a higher confidence towards classifying an image as an example of $k^{th}$ class, for $W_{ik} < 0$ increase in pixel brightness will respectively decrease the classification confidence.

Representation of all MNIST classes in a form of a pixel importance map towards classifying each image as an instance of $i^{th}$ class is shown on figure 1. The presented illustrations can be thought of as some generic neural network digit representation and in most cases these representations can be recognized as rough shapes of actual handwritten digits.

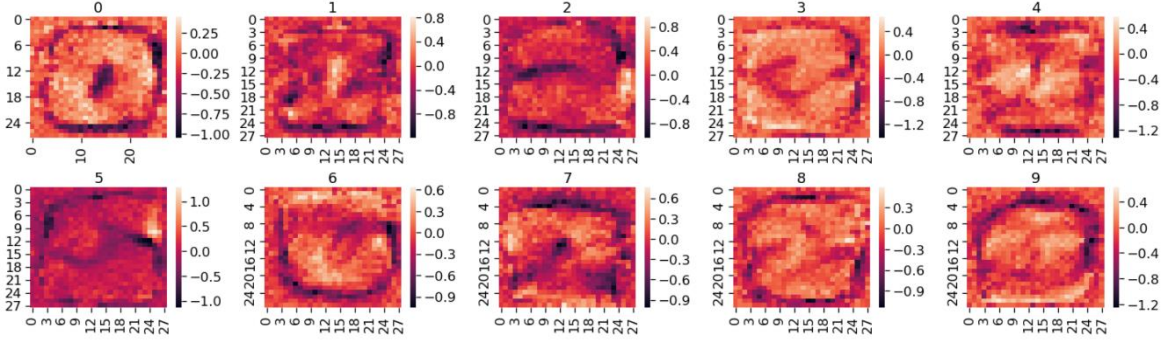

Figure 1 – Pixel importance map towards classifying each image as an instance of $i^{th}$ class

The result of a subtraction $W_{i9} - W_{i4}$, $\forall i$ is shown on figure 2. By building such a weight difference we can easily see that the most important region for maximizing images inference towards the class of nines is below the shape of four (shown in light tones). It makes sense as bottom line is a piece that must be present in a nice and must not be present in a four. Regions that minimize digit's classification probability towards nine lie outside the shape of a digit in the top right and left corners and in the middle left, likely hinting that nines should have rounded edges and fours should not. Such an intuitive pixel importance map representation will lie a foundation of the first adversarial attack algorithm.

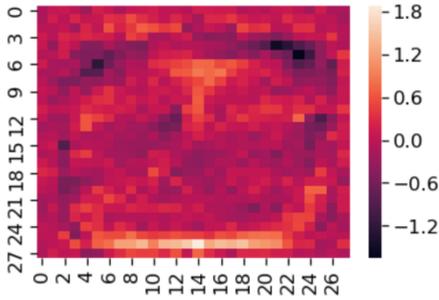

Figure 2 – Pixel importance difference for classes of 9 and 4

In order to make any conclusions about algorithm's efficiency we need to introduce a quality metric. To estimate image quality loss $L_\infty$-norm has been used in [8], that is the largest deviation of a pixel brightness over an entire image. For images, whose pixel values are bound in the range [0,255] deviations up to 15 points were permitted. However, such a metric allows to generate nearly unrecognizable (when compared to the source) images, which is not something we are up to. So, a metric that is highly correlated with a human perception is needed. The best results can be obtained by using one of the following metrics: MAE (Mean Absolute Error), PSNR (Peak Signal to Noise Ratio), SSIM (Structural Similarity Index). The first two are easy to compute and are frequently used, but they do not take human vision features into account. SSIM metrics has been introduced as an improvement on top of MAE and PSNR. As it has been shown in [21], it is possible to craft images with different distortions namely adding noise, changing brightness/contrast, blurring the image, that will get the same MAE score, while the images are vastly different and given dissimilar scores by the human observers. SSIM score has no such a deficiency and outputs scores that are highly correlated with a human vision. SSIM metric values lie in range $[-1,1]$. The maximum value signifies that images are identical. PSNR is also a more efficient than MAE, but its scores have no upper bound, meaning that for identical images we will get an infinite score, thus making nearly impossible to generate a consistent summary of adversarial attack. Hence, SSIM metric is to be used for the algorithm's quality estimation.

Let's denote neural network attack problem statement. The output is presented as a probability distribution of handwritten digit classes $z$. Consider that the network predicts image as an instance of class $S \in \{0,1,2,\ldots,9\}$ if $z_S = \max_{k=\overline{1,K}} z_k$. By changing some pixels' brightness, we want to change neural network prediction to $T \in \{0,1,2,\ldots,9\}$, $T \neq S$, i.e. $\hat{z}_T = \max_{k=\overline{1,K}} \hat{z}_k$, where $\hat{z} = \sigma((x + \Delta x) \cdot W + b)$. Also we enforce image correctness by clamping brightness values into a range $x_i \in [0,1]$, $i = \overline{1,784}$.

**Single algorithm step pseudocode**:
attack_step (*image*, *source_weights*, *target_weights*, $\alpha \in \{0; 0.5; 1\}$, $min\_difference > 0$, $step > 0$*)*
  for each point $i$ in the *image*
    find corresponding weights $W_S$ and $W_T$ in weight matrices for source $S$ and target $T$ classes
  let $delta = \alpha W_T - (1-\alpha) W_S$
  if $|delta| > min\_difference$, then
    $x = x + step \cdot delta$
  if $x < 0$, then $x = 0$
  if $x > 0$, then $x = 1$.

Where *image* is an image obtained on the previous algorithm step or the source one if this is the first algorithm step; *source_weights* is the trained network weight matrix for original image label; *target_weights* is the trained network weight matrix for a class, towards which we want to change the prediction; *min_difference* is the minimal difference between classes' weight matrices for a pixel to be an attack target; *step* signifies pixel brightness change on the current iteration; $\alpha \in \{0; 0.5; 1\}$ defines the algorithm modification, the algorithm step is repeated *max_steps* times. Recommended values for the described parameter values are to follow.

As it was remarked above, by the constraints on the target class $j$, adversarial attacks are divided into two subtypes:

– if the goal is to assign to an image a specific class $j$ instead of class $k$, then such an attack is called targeted. This type of attack can be accomplished via the described algorithm with $\alpha = 0.5$ or $\alpha = 1$. With that, it is said that the algorithm has succeeded to attack an image only if the algorithm has been able to change neural network's predicted class into digit $j$ in a finite number of steps, and has failed in all other cases;

– if the goal is to reassign classification of an image of a class $k$, to any different one $j \neq k$, then the attack is called non-targeted. The algorithm parameter $\alpha = 0$ can be used to perform such an attack. The result is a success if an incorrect classification has been achieved in a finite number of steps.

## 4 EXPERIMENTS

Here we will present several of most interesting attack cases. $7 \to 2$ is the first attack to be performed with a goal to force a neural network to misclassify 7 as 2. Adversarial attack results are shown on fig. 3. Algorithm parameters:

$\alpha = 0.5$, *min_difference* = 0.0, *step* = 0.02, *max_steps*=10. The image with perturbations that were enough to trick the network into thinking that it has been presented a "2" digit is highlighted in green. As the image appears to be visually indistinct from the source (which is confirmed by a high SSIM metric of 0.931), the image difference has been visualized below, where white regions mean that no changes have been applied in that point, red regions signify source image brightness increase, the blues mean brightness decrease.

It should be considered, that if all image pixels are being attacked by bearing in mind only the sign of the attack difference (as it is done in [6]), then the attack will still be successful. However, in such a case image noise can be viewed easily. A generic comparison with fast adversarial attack methods will be presented in a later section of the paper.

Let's visualize influence of *min_difference* attack parameter onto the attack result. The algorithm step has been increased for the effect to be more pronounced. In our case 99% of pixel weights lie in range $[-1;1]$, so the difference by modulo is within $[0;2]$, that's the range for the *min_difference* parameter. Fig. 4 has the results of targeted attack $7 \to 2$ displayed with algorithm parameters: $\alpha = 0.5$, *min_difference* = 0.65, *step* = 0.25, *max_steps* = 10. As it turns out, it has been enough to modify pixel brightness of only 3 source image points for the attack to succeed. SSIM value of 0.926 has been achieved. Experiments akin to this one have been performed on a set of images and all with a success. However, it has been noted, that an algorithm has an interesting feature, where in some cases it leads the attack not directly to the target class, but through some intermediate class instead.

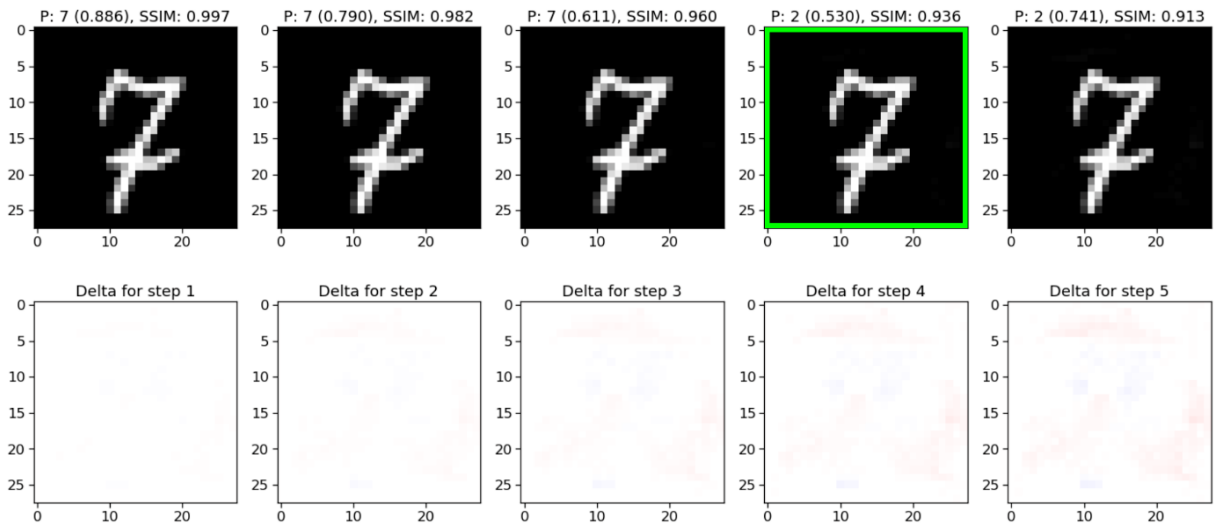

Figure 3 – 7→2 attack with low min_difference. Algorithm parameters: $\alpha = 0.5$, *min_difference* = 0.0, *step* = 0.02, max_$steps$ = 10

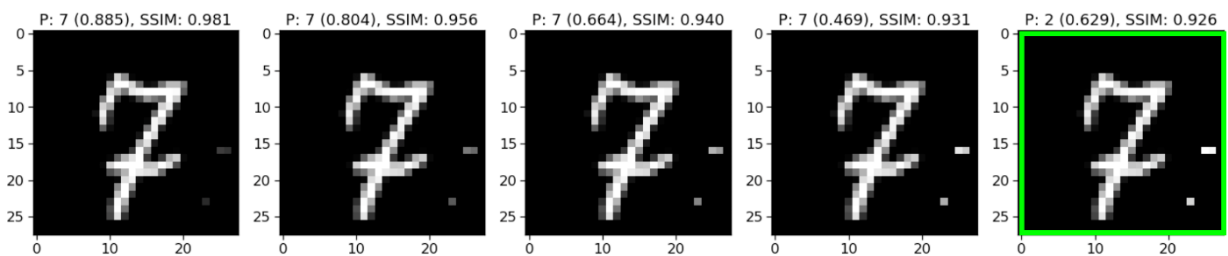

Figure 4 – 7→2 attack with high min_difference. Algorithm parameters: $\alpha = 0.5$, *min_difference* = 0.65, *step* = 0.25, *max_steps* = 10

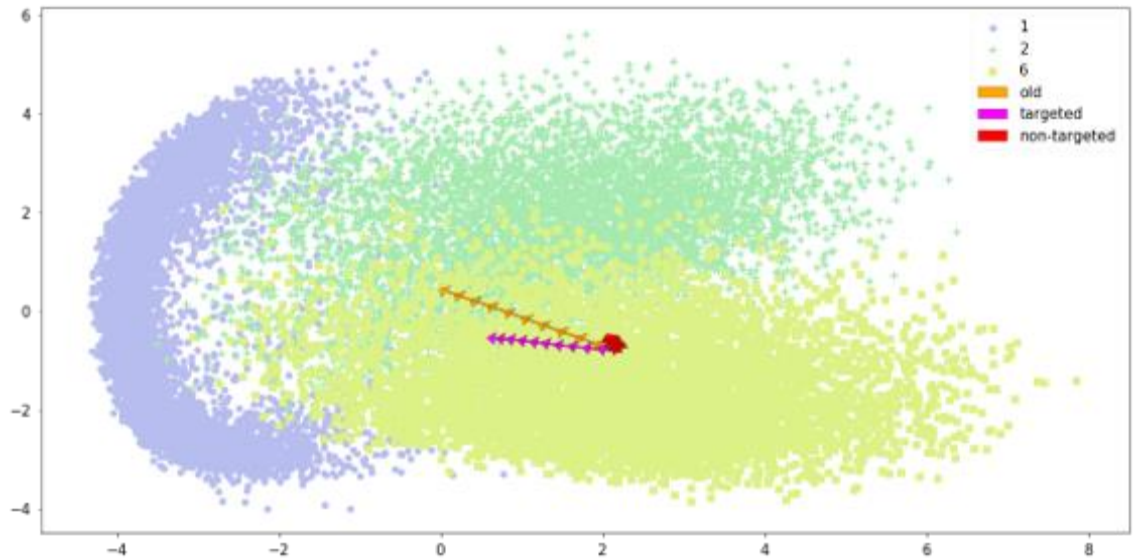

Figure 5 – 6→1 attack trajectory as projected onto a surface using PCA: targeted attack with $\alpha = 0.5$ (orange), targeted with $\alpha = 1$ (purple), non-targeted with $\alpha = 0$ (red)

For example, for the targeted attack 6 into 1 during the first algorithm steps we have 6 misclassified as 2. In a case when such an intermediate class appears during the attack, perceptual image quality is degraded (for this example we got SSIM = 0.635).

The problem's roots have been investigated by utilizing PCA (Principal Component Analysis). After having visualized scatted plot for points of classes 2, 6 and 1 it has been noticed, that intercluster distance for digits 2 and 6 is a lot lower that the distance for 6 and 1. And as is shown on fig. 6 (for α=0.5), the difference vector between the target and source classes passes through the field of twos.

For targeted attack such issues can be avoided by using another algorithm modification with parameter $\alpha = 1$. Thus, by applying such a modification for the targeted attack we will strive to maximize output of the target neuron when compared to others. We can minimize source neuron output in respect to the other ones to accomplish non-targeted attack using an additional algorithm modification with $\alpha = 0$.

Much better modified image quality can be obtained by the virtue of such algorithm modifications (for example, for the same 6→1 attack SSIM score has risen to 0.780). Fig. 5 has trajectories of the source image 6 while being a subject to modification by the original algorithm and both its variations (targeted and non-targeted).

## 5 RESULTS

So far, we have presented two images for an attack: one has needed almost no changes for the attack to succeed, another has visible changes introduced to be misclassified yet it is still correctly recognized by the human observers. In order to compare our algorithm against other fast adversarial attack algorithm we need to perform a generalized attack analysis. By launching targeted attack for each pair of source and target classes, a success rate heatmap has been drawn (fig. 6a). Source classes are shown on the left, the target classes in the bottom.

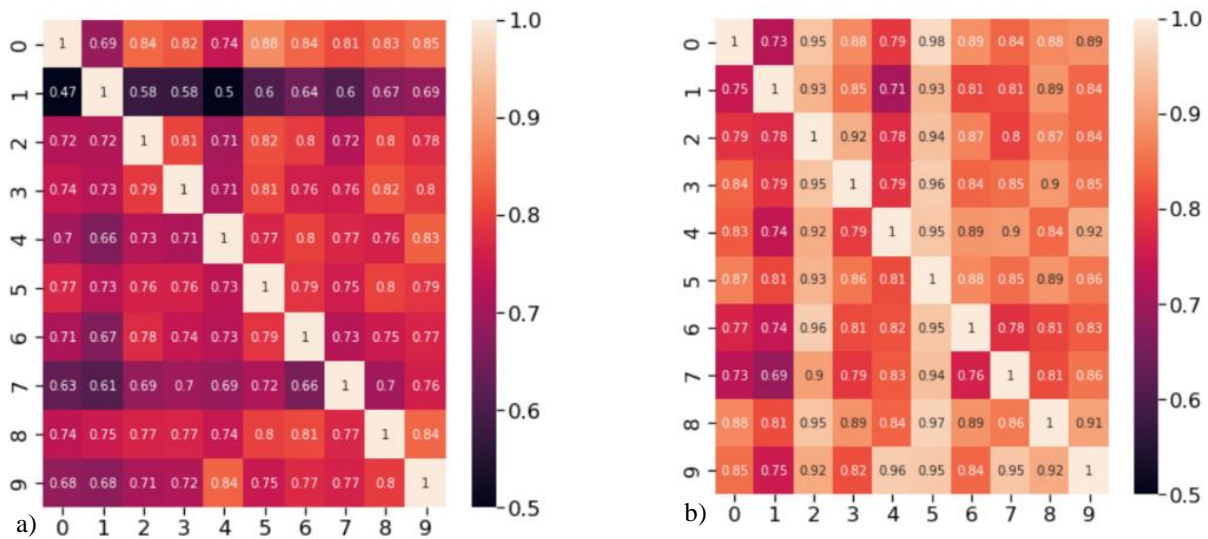

Figure 6 – SSIM metric values heatmap for each source, target attack pair: a) with fixed adversarial attack scoping; b) with automatically selected adversarial attack scoping.

Heatmap elements are SSIM values averaged across all the attacks for a given source, target pair. Mean quality over the whole test dataset is 0.76 – such images after attack will still be correctly classified by a human. Top score has been achieved for an attack of similar digits i.e. 8→9, 9→8, 0→8, 3→8. The highest quality degradation was for attack 1→0. This can be explained by the fact that the vital region for zero is a black hole in the middle, which gets usually overlapped by a white bar of a one digit. Should be noted, that 0→1 attack requires much fewer image modifications then the one in opposite direction, which is proved by comparing SSIM value (higher by 0.21). As attacks have been built on real test set samples opposed to the generic digit silhouettes which got learnt by the neural network, the heatmap SSIM values lack symmetry. Lower average image quality loss can be attained by employing a stricter parameter selection algorithm. While fig. 6 has losses computed for a low empirically chosen *min_difference*=0.5 value, by selecting the best value from range [0.0, 1.0] an increase of SSIM to 0.87 score has been observed (fig. 6b). A higher average SSIM score 0.93 has been reached for the non-targeted attack case, which means that source and target images are nearly impossible to distinguish with a naked eye. As previously, digits 0, 8 were the easiest ones to attack, a class of ones has proved to be the most problematic (fig. 7).

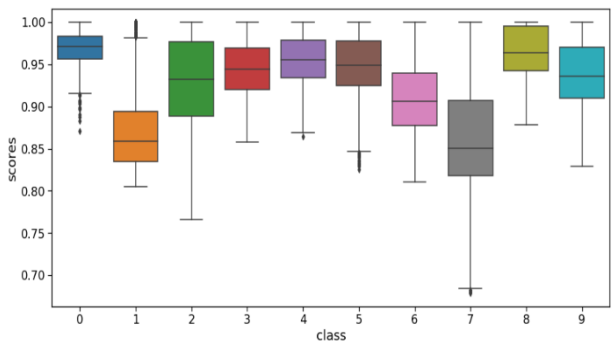

Figure 7 – Low non-targeted attack SSIM score deviation. Algorithm is good at attacking different images of each class.

If image quality loss deviation is high for different images of a certain class (i.e. some images are easy to attack, while others not), then algorithm is inefficient (as it can only change digits that look similar to several classes); if, conversely, the deviation is small, then the algorithm is efficient. So, by checking our targeted attack box-plot against the above-described thought we have come to a conclusion, that our algorithm is efficient at attacking different classes.

SSIM plots, with respect to *min_difference* parameter value, have allowed to make a conclusion about the fact that each class has a tendency of an image quality rise jointly with *min_difference* increase up to 0.9 point, such a trend is especially noticeable for the class of nines (fig. 8).

Specifically, the human perceived image quality loss will be substantially lower in case of a strong change of several pixels, then when all image points are slightly modified. Taking this feature into account is the thing that makes our algorithm standout among all other known methods in the literature.

Among the fast gradient methods, the most efficient algorithm is I-FGM with $L_2$ norm loss [7]. An attack for each source, target pair has been conducted by following the above described procedure for the case without *min_difference* selection. Algorithm has been successful on all test images, but has achieved a lower SSIM score of 0.83.

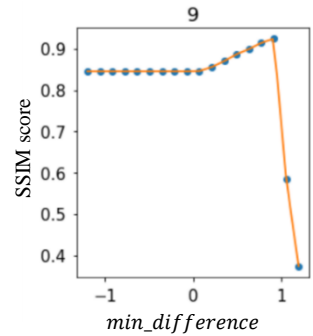

Figure 8 – Increase of SSIM score when attack scope becomes smaller. The scope that is too small makes the attack impossible.

Next, the question of adversarial image transfer has been considered. We want to perform the so-called black box attack, when we don't have any knowledge about network's weights or architecture, the only allowed operation is to query neural network prediction engine by submitting images to it. The attack will be performed by using the above described logistic regression architecture, then an attempt to transfer each image to a 5-layer unknown neural network will be made.

For the results reproducibility neural network architecture is to follow, yet this knowledge has not been used in any way during the attack phase. The neural network has a 5-layer fully-connected architecture with layer sizes of 200, 100, 60, 30, i.e. 4 hidden, one output with 10 neurons and one input with 784. As a mean of regularizing the network Batch Normalization has been applied after the first layer, Dropout after the second one. ReLU has been used as an activation function for all layers but the last one, where we have switched to a Softmax function instead. After 100 epochs of training using Adam optimization algorithm, training set accuracy has reached 98.65%, the test one 98.51%.

By the above-described procedure an average probability of 33% successfully transferred images has been achieved. Interesting to note, that in many cases images that were difficult to attack for the original network have seen a higher transfer rates than the ones needed only minor image changes. For instance, it has been possible to successfully transfer 87% of 1→0 attack images, which have been one of the most challenging ones, but only 14% 9→7 attack images.

Let's follow along the 0→8 attack procedure. Each step will have the predicted digit with its probability shown for the attacked single-layer classifier (SL) and 5-layer fully-connected network (FC5) (fig. 9). It should be observed that after there were enough changes to cheat the original network, it has been necessary to make 4 more steps to deceive the 5-layer one. This means that while the two networks have a similar decision boundary yet each one has it biased with respect to another one. Funnily, the changes introduced resemble a doorbell with a bowknot, no similarity with a sample of eight is observed.

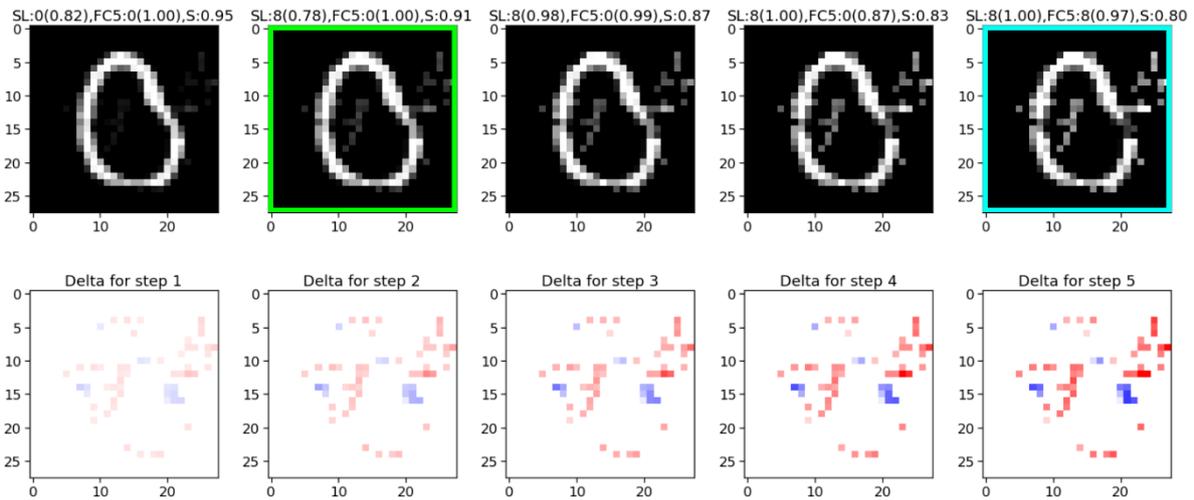

Figure 9 – 0→8 transfer attack. Images are transferred to a fully-connected 5-layer network. Algorithm parameters:
$\alpha = 1.0$, $min\_difference = 0.2$, $step = 0.4$, $max\_steps = 10$

Considering the above-described thoughts, a generalized targeted attack with neural network transfer has been conducted once again. This way, after having performed a successful attack on the source network, 4 more algorithm steps are made. After that target network is queried only once for each *min_difference* value in a range [0.0,1.2] with step 0.1, this procedure yields 12 images for each *min_difference*, the best one of which is selected based on target (transfer) network score. This complication is done to minimize transferred network call count as the access to the network can be slow or partially restricted. This procedure makes it possible to transfer 91% of adversarial images with a minimal image quality loss.

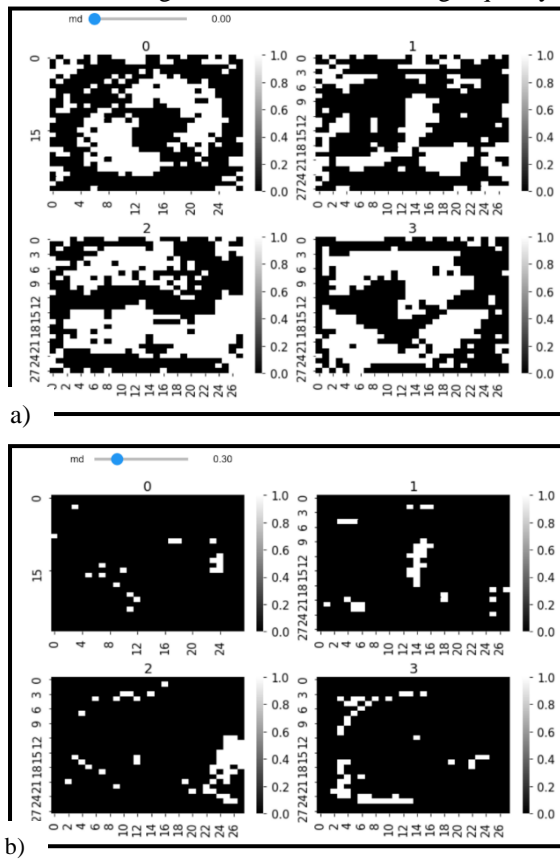

Figure 10 – An example of the generated importance map.
a) $min\_difference = 0.0$; b) $min\_difference = 0.3$

Furthermore, for most of source, target class pairs the attack transfer has been successfully performed for 100% images, except for the class of sixes.

Having performed a successful black-box attack using adversarial image transfer from a single-layer network to a vastly different 5-layer network, it has been decided to try to tackle a real-world attack scenario. As has been mentioned above neural networks are getting traction in the domain of robotics, CCTV systems and autonomous vehicles, this is where adversarial attacks are the most dangerous. However, in such cases it might be hard to perform a "man-in-the-middle" type of attack, where an attacker can get into the stream between a camera and corresponding processing system, so we need to make adversarial image sustain artefacts and noise introduced by the camera.

For the experiment we are going to use digits that are written on a white paper with a black pen, shots are to be taken with an ordinary smartphone camera. In order to make a guarantee of digits being correctly classified by the neural network, we need to perform the same conversion steps as done on the images of the MNIST dataset. Namely, we need to 1) convert color image into grayscale; 2) compute a center-of-mass of the pixels for a given digit and crop around that region a $28x28$ box; 3) invert colors: as we need to have a black background and white ink. Thus, a set of recognitions has been performed, the network has been successful at recognizing digits written under various angles and by different people.

Next, we need to perform an attack. To restrict perturbations done to the image let's assume that writing with a pen is the only allowed operation, no parts of a digit can be erased. Also, as it has been mentioned previously, targeted attack case is more challenging than untargeted, so we will focus only on a targeted attack. Given the above-described constraints, an algorithm constructing binary importance map has been implemented. With a minimum value of $min\_difference = 0.0$, generally the shape of the digit is well visible (fig. 10a). However, by scoping importance region with a value of $min\_difference = 0.3$, clouds of dots with high network weight or high importance start appearing (fig. 10b).

With that a set of attacks has been performed. Next a few examples selected at random are presented. In a case of 4 → 9 attack, initially we had 4 recognized with a probability of 80%, after underlining the digit we get 9 with confidence 42% (highest confidence for the given image), as shown on fig. 11a. 0 → 5 attack has been performed by adding a quote in the top-right corner (fig 11b). Interestingly, while 0 was recognized with 80% confidence, 5 has a confidence of 100%. Lastly, two attacks on 9 are shown: 9 → 6 by drawing a white blob near the digit, 9 → 5 by writing a comma (fig. 11c). In all the cases an attack region to be drawn with a pen has been selected by consulting the binary importance map. It is important to note, that generally it has been possible to perform the attack outside a digit's bounding box, thus the shape is easily recognized by the human, additionally, no

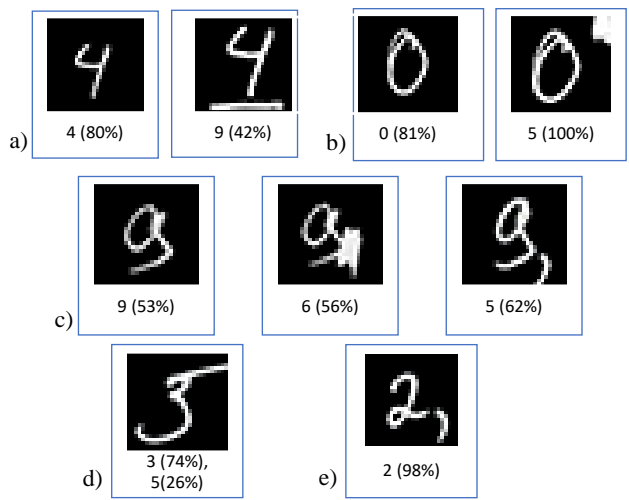

Figure 11 – Examples of conducting attacks in real-world
a-c) Attacks performed with the help of the designed importance map are successful
d-e) Random perturbations without knowing vulnerable regions have no effect distracting noise has been added to the image. However, the question arises whether any random perturbation can full the network. This question has already been answered in [3] for adversarial images fed into the network directly, where the answer was negative – only for a small set of images it has been possible to change assigned class by the neural network. Does the same holds for a physical-world type of attack? It does. For example, digit 3 with an extended line is similar to 5 for a human, so it is for a neural network (fig. 11d), adding comma to a digit 2 has not helped it to be recognized as 5 (fig. 11e) as in the case of 9 → 5 attack.

## 6 DISCUSSION

A neural-network adversarial attack algorithm based on a scoping idea has been presented, with interesting features such as an ability to draw binary importance map with region highlighting for a targeted attack for a specific class, which allows for the first time to perform a MNIST handwritten digit dataset attack using an ordinary pen or pencil. By varying area of the attack using the introduced *min_difference* parameter it has become possible to improve effectiveness (in terms of minimizing structural image similarity loss during the attack) of existing fast gradient-based adversarial attack methods and to perform a successful adversarial attack transfer with minimum number of transferred network calls, thus enabling efficient attacks against yet unknown systems.

The presented adversarial attack analysis scheme will ease early neural-network-based systems' security diagnostics and will boost further research in the direction of better understanding of inner workings of neural network's inner inference mechanisms.

## CONCLUSIONS

Three fast adversarial attack algorithm modifications for logistic regression have been presented in the paper: two for performing targeted and one for non-targeted attack. It has been shown that by scoping adversarial attack it is possible to lower quality loss during the attack when compared to other fast gradient-based adversarial attack algorithms and to generate binary adversarial maps for further efficient attack transfer into real world using pen or pencil. The relevance of the work is explained by a growing use of neural networks in security critical areas without prior analysis of the neural network's durability against attacks.

The scientific novelty of obtained results is that for the first time adversarial attack algorithm has been built upon the attack scoping idea. The presented algorithm can automatically select region size for adversarial attack thus yielding images that have fewer structural changes than other gradient-based algorithms. Adversarial attack scoping has an interesting feature of improving adversarial attack transfer across networks of different architectures and into the real world.

The practical significance of the obtained results is that an early neural network vulnerability diagnostic can be performed by utilizing the proposed algorithms and image quality loss analysis system, which is a pivotal point towards a safer practical neural network use.

## ACKNOWLEDGMENTS

The authors would like to thank Doctor of Sciences in Physics and Mathematics, Associate Professor of the Department of Data Analysis and Artificial Intelligence of the National Research University "High School of Economics" Gromov Vasilii Aleksandrovich for support and a fruitful paper discussion.